\documentclass[10pt,twocolumn,letterpaper]{article}

\usepackage{wacv}              
\usepackage{graphicx}
\usepackage{amsmath}
\usepackage{amssymb}
\usepackage{booktabs}
\usepackage{url}
\usepackage{multicol}
\usepackage{balance}
\usepackage{multirow}
\usepackage{makecell}
\usepackage{float}

\usepackage[pagebackref,breaklinks,colorlinks]{hyperref}

\usepackage[capitalize]{cleveref}
\crefname{section}{Sec.}{Secs.}
\Crefname{section}{Section}{Sections}
\Crefname{table}{Table}{Tables}
\crefname{table}{Tab.}{Tabs.}

\def\wacvPaperID{1162} 
\def\confName{WACV}
\def\confYear{2025}

\begin{document}

\title{Enhancing Skin Disease Diagnosis: Interpretable Visual Concept Discovery\\ with SAM\thanks{$\dag$ indicates equal contribution, $\ddag$ indicates the corresponding author \url{zding1@tulane.edu}. This work is partially supported by the Louisiana Board of Regents Support Fund (LEQSF(2022-25)-RD-A-22) and NIH 2U19 AG055373-06A1.}}

\author{Xin Hu$^{1,\dag}$, Janet Wang$^{1,\dag}$, Jihun Hamm$^1$, Rie R Yotsu$^2$, Zhengming Ding$^{1,\ddag}$\\
$^1$Department of Computer Science, Tulane University\\
$^2$Department of Tropical Medicine, Tulane University\\
}

\maketitle

\begin{abstract}
Current AI-assisted skin image diagnosis has achieved dermatologist-level performance in classifying skin cancer, driven by rapid advancements in deep learning architectures. However, unlike traditional vision tasks, skin images in general present unique challenges due to the limited availability of well-annotated datasets, complex variations in conditions, and the necessity for detailed interpretations to ensure patient safety. Previous segmentation methods have sought to reduce image noise and enhance diagnostic performance, but these techniques require fine-grained, pixel-level ground truth masks for training. In contrast, with the rise of foundation models, the Segment Anything Model (SAM) has been introduced to facilitate promptable segmentation, enabling the automation of the segmentation process with simple yet effective prompts. Efforts applying SAM predominantly focus on dermatoscopy images, which present more easily identifiable lesion boundaries than clinical photos taken with smartphones. This limitation constrains the practicality of these approaches to real-world applications. To overcome the challenges posed by noisy clinical photos acquired via non-standardized protocols and to improve diagnostic accessibility, we propose a novel Cross-Attentive Fusion framework for interpretable skin lesion diagnosis. Our method leverages SAM to generate visual concepts for skin diseases using prompts, integrating local visual concepts with global image features to enhance model performance. Extensive evaluation on two skin disease datasets demonstrates our proposed method's effectiveness on lesion diagnosis and interpretability. 
\end{abstract}


\section{Introduction}

\begin{figure}
    \centering
    \includegraphics[width=.95\linewidth]{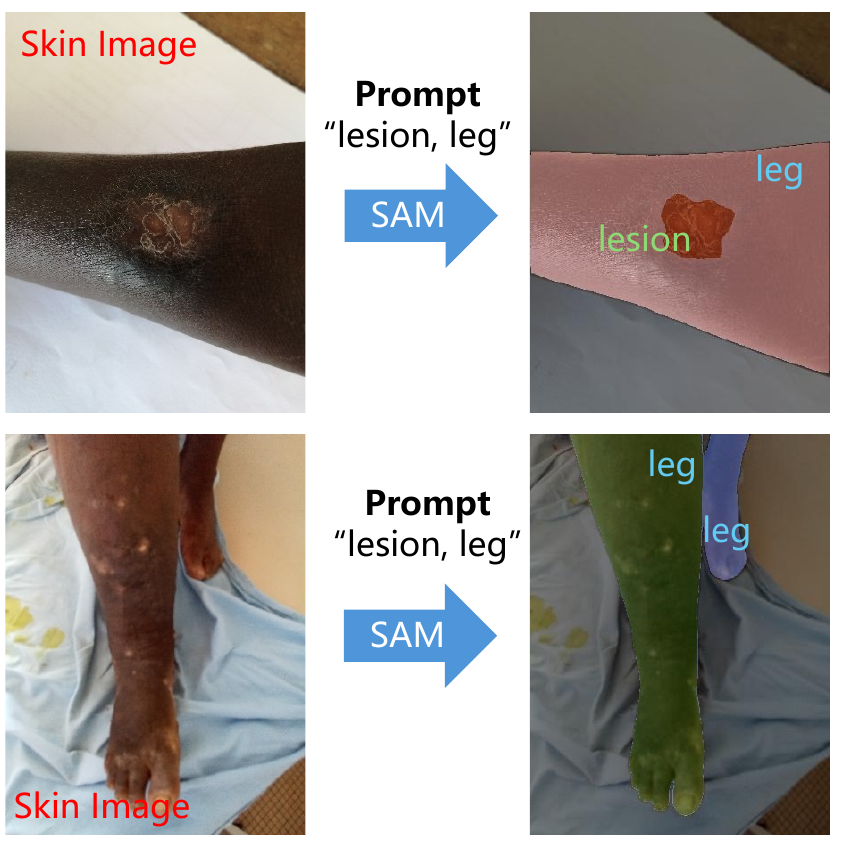}\vspace{-2mm}
    \caption{Generated masks on two skin samples with prompt ``lesion'' and ``leg''. The first row shows a ``Buruli ulcer'' image, in which the lesion part is clear to be detected as the visual concept, while in the second row - ``Mycetoma'', the lesion boundary is ambiguous to be recognized.}\vspace{-3mm}
    \label{SAM_mask}
\end{figure}

Current AI-assisted diagnostic systems demonstrate expert-level capability in classifying skin cancers, which are often identified visually \cite{Esteva2017DermatologistlevelCO, Liu2020-vx, Brinkerarticle, tschandl2020human, soenksen2021using, celebi2019dermoscopy, wang2024achieving}. Given that early and accurate diagnosis is important in improving treatment outcomes, these systems can significantly contribute to teledermatology as diagnostic and decision-support tools \cite{freeman2020algorithm}. By utilizing photos captured from portable devices like smartphones, these methods promote diagnostic accessibility in rural areas \cite{Coustasse2019UseOT}. However, such systems are susceptible to under-diagnosis due to non-standardized acquisition environments and protocols. Therefore, the model's ability to accurately recognize the region of interest (ROI) within noisy backgrounds is crucial for precise diagnosis.

While image segmentation techniques such as MaskRCNN \cite{he2017mask}, DeepLab \cite{chen2017deeplab}, and Panoptic Segmentation \cite{kirillov2019panoptic} can be employed to localize ROIs and enhance diagnostic accuracy, they rely on fine-grained bounding box or pixel-wise semantic annotations. To acquire such annotations from manual labeling is labor-intensive and impractical, especially when dealing with noisy images. The Segment Anything Model (SAM) \cite{kirillov2023segment} can serve as a potential solution, capable of automating the generation of two-level masks: one for the region of the body and another for the lesion. These masks also enable the explainable investigation of different features in the diagnostic results.

Two works have explored SAM to locate skin lesions and automate the segmentation \cite{hu2023skinsam, himel2024skin}. Both studies primarily focus on the well-curated HAM10000 Dataset, which includes 10,015 verified dermoscopy images and groundtruth segmentation masks \cite{DVN/DBW86T_2018, tschandl2020human}. In this dataset, examples are abundant, lesion boundaries are clear, and background noise is minimized. In contrast, clinical photos collected in real-world applications present greater difficulty. The contrast between lesions and backgrounds in these photos is often less distinct due to factors such as (1) a wide variance in lesion presentation, including differences in shape, texture, color, and location, and (2) inconsistent image quality caused by varying lighting, distance, angles, environments, or image resolutions. These variations make it difficult for the model to accurately identify the ROI from the background, limiting the practicality of existing methods. Moreover, clinical photos of skin conditions are often limited in size. Therefore, the potential application of SAM in small datasets of clinical photos remains under-explored. Figure \ref{SAM_mask} shows results on example images using SAM.

In this work, we propose an innovative framework to locate the region of interest and diagnose skin diseases in the clinical photos. Empowered by SAM, our framework generates two-level masks to enhance the accuracy of AI diagnoses and provide robust support for remote medical personnel, enhancing diagnostic reliability and transparency. To sum up, we highlight our contributions as follows: \vspace{-1mm}
\begin{itemize}
\item We leverage the capabilities of a foundational segmentation AI model to generate two-level masks over the body and lesions with prompts, which delineates the regions of interest and reveals potential visual concepts essential for skin diagnosis.\vspace{-2mm}
\item We introduce a novel Cross-Attentive Fusion framework that integrates knowledge from global image features and local visual concepts to enhance the model's diagnostic accuracy. This framework also facilitates interpretable analysis of visual concepts crucial for accurate diagnosis.\vspace{-2mm}
\item We comprehensively evaluate our method on two distinct skin image datasets of challenging clinical photos. Through detailed interpretation and analysis, our method demonstrates robustness and effectiveness in real-world scenarios, elevating confidence in its diagnostic capabilities.
\end{itemize}
\label{sec:intro}

\section{Related Works}
\label{sec:related_works}

\subsection{Visual Segment Generation}
Image segmentation techniques such as MaskRCNN \cite{he2017mask}, DeepLab \cite{chen2017deeplab}, and Panoptic Segmentation \cite{kirillov2019panoptic} are effective for localizing regions of interest and enhancing model accuracy, but they typically require fine-grained bounding box or pixel-wise semantic annotations. With a growing focus on scalability, pre-trained foundational models have gained prominence in machine learning, serving as robust starting points for various downstream tasks \cite{thrun1995learning}. Responding to this trend, the Segment Anything Model (SAM) has emerged, pushing image segmentation into the realm of foundational models.

\begin{figure*}[t]
    \centering
    \includegraphics[width=1\linewidth]{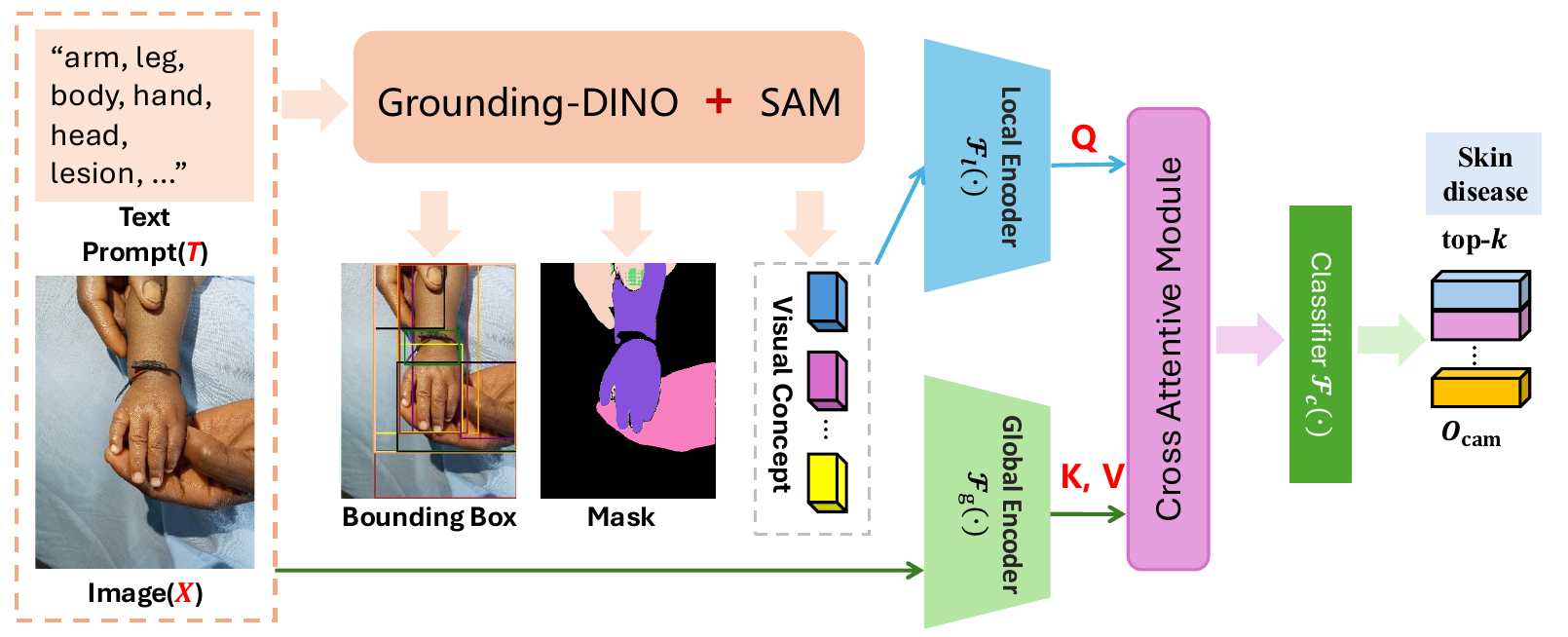}\vspace{-2mm}
    \caption{Overall framework of our proposed model, where we use Grounding-DINO and SAM to extract visual concepts, bounding boxes, and segmentation masks. The local encoder $\mathcal{F}_l(\cdot)$ converts visual concepts to local tokens and sets them as ``query'' prompts to trigger the salient area of the encoded global image with cross attentive module. The classifier $\mathcal{F}_c(\cdot)$ transfers the latent features to CAM for classification and interpretation for the decision-making process.}\vspace{-4mm}
    \label{framework}
\end{figure*}

SAM is a promptable segmentation model pre-trained on SA-1B, a vast dataset containing over 1 billion masks derived from 11 million licensed and privacy-preserving images. This extensive training ensures strong generalization across diverse data distributions. SAM supports flexible prompts such as single points, sets of points, bounding boxes, or text. Its architecture is elegantly simple yet effective: a powerful image encoder computes an image embedding, a prompt encoder embeds prompts, and these inputs are fused in a lightweight mask decoder to predict segmentation masks. Specifically, SAM's image encoder utilizes a Masked Autoencoder (MAE) \cite{he2022masked} pre-trained on a Vision Transformer (ViT) \cite{dosovitskiy2020image}. This adaptation allows it to handle high-resolution inputs while capturing fine-grained details and long-range dependencies. The prompt encoder converts prompts into fixed-length embeddings that capture semantic meanings. These embeddings are combined with image encoder outputs to generate a set of feature maps used by the mask decoder to produce segmentation masks.

\subsection{Skin Lesion Analysis}

Skin lesion images are typically captured in two forms: dermatoscopy images and clinical photos. Dermoscopy images are close-up views of pigmented skin lesions captured by using professional microscopy, a dermatoscope. These images focus on the lesion, typically excluding any background, resulting in a uniform and consistent visual presentation. Clinical photos, on the other hand, are often taken with portable devices like smartphones and contain more noisy backgrounds. Extensive research has explored the use of deep Convolutional Neural Networks (CNNs) and Generative Adversarial Networks (GANs) \cite{goodfellow2020generative, narayanan2023multi} for skin lesion segmentation to standardize lesion variations and enhance diagnostic accuracy \cite{ozturk2020skin, xie2020skin, hasan2020dsnet, zafar2020skin, lei2020skin, innani2023generative}. CNN-based methods for skin lesion segmentation typically employ supervised learning with large labeled datasets to extract spatial features and semantic maps from images. In contrast, GAN-based approaches address data scarcity through unsupervised learning but still require fine-grained groundtruth masks, often unavailable in real-world applications.

With the emergence of large foundation models, there has been an effort to utilize the Segment Anything Model (SAM) for dermatoscopy images, using simple prompts \cite{hu2023skinsam, himel2024skin}. Specifically, \cite{hu2023skinsam} fine-tuned SAM on the HAM10000 Dataset with fine-grained groundtruth masks, achieving state-of-the-art segmentation performance with the guidance of prompts. \cite{himel2024skin} utilized SAM to segment cancerous areas in HAM10000 images before feeding the segmented images to a pre-trained Vision Transformer (ViT) model for classification. Their results demonstrated ViT's superiority over traditional architectures. Despite these advancements, adapting SAM to clinical data captured by portable devices poses challenges due to limited dataset sizes for fine-tuning and the complexity of lesion manifestations for off-the-shelf SAM applications \cite{zhang2024improving}. Therefore, the potential of SAM in low-data scenarios and with clinical photos remains to be fully explored.

\section{Methods}
\label{sec:methods}

\subsection{Preliminary and Motivation}
Our objective is to develop an accurate and interpretable diagnostic model for complex image data. Given a batch of skin disease images, \(\mathbf{X} = \left\{\mathbf{x}_{1}, \mathbf{x}_{2}, \mathbf{x}_{3}, \ldots, \mathbf{x}_{i}, \ldots, x_{N} \right\}\), where \(N\) is the total number of images, the model aims to predict the most likely skin disease for each image, represented as \(\{\mathbf{y}_{i}\}_{i=1}^{N}\). Here, \(\mathbf{y}_{i} \in \mathbb{R}^{C}\), with \(C\) representing the total number of condition categories, and \(\mathbf{y}_{i,c}\) the presence of skin disease category \(c\) in \(\mathbf{x}_{i}\).

Considering the skin images are usually captured from smartphones and tablets in the field, they lack standardization and often contain confounding background noise. The ability to easily discern the region of interest (ROI) or lesion in a poor-quality or noisy image is crucial to accurate diagnosis. In this study, we will explore the Segment Anything Model (SAM) \cite{kirillov2023segment} to detect visual concepts with specific prompts, then develop a cross-attentive fusion model by leveraging the global image information and local visual cues to improve the skin disease diagnosis performance.

\subsection{Visual Concepts Identification}

Due to the often complicated background noise, end-to-end deep learning is not sufficient to fully extract informative features from raw images with disease labels. Though SAM can capture local visual concepts of lesions, the original SAM needs bounding boxes or coarse masks as input and might produce features containing background noise. Thus, we modify variants of SAM in public repository \cite{track2023} which utilizes Grounding DINO\cite{liu2023grounding}, to obtain visual concepts and provide related bounding boxes.

As shown in Figure \ref{framework}, we use keywords describing human body parts and ``lesion" as text prompts and input them with the original image to the SAM variant to generate bounding boxes, visual concepts, and segmentation masks. Figure \ref{SAM_mask} lists some detected visual concepts. It is noted that the text prompt ``lesion'' might correctly localize the pathological area for certain skin diseases whose lesion boundaries are clear, but it is challenging for other complicated conditions, such as ``Mycetoma''. Still, these captured visual concepts can offer extra information and help us identify the relevant segments of the body or the lesion. 

We define local visual concepts as $\{ \mathbf{v}_{1}, \mathbf{v}_{2}, \ldots \} \in \mathbf{V}$. By focusing on these localized visual elements, our model can diagnose skin diseases more accurately and effectively. Additionally, given the visual concepts identified by SAM, we can enhance the interpretability of our model by pinpointing the specific local patterns that lead to a prediction. The framework provides valuable insights into the decision-making process, thereby increasing its transparency and trustworthiness as a diagnostic system. This dual focus on accuracy and interpretability is crucial for developing a reliable diagnostic tool that can assist healthcare professionals in making informed decisions.

\subsection{Cross-Attentive Fusion Model}

Due to the unreliability of certain visual concepts from SAM (Figure \ref{SAM_mask}), we concurrently leverage both global image features and local visual concepts. As illustrated in Figure \ref{framework}, our proposed dual-branch framework is designed to harness the complementary strengths of these two sources of information. This approach allows for a more robust and comprehensive understanding of the diseases by integrating detailed local visual concepts with overarching global image features, thereby improving the overall accuracy and reliability of the analysis.

Firstly, we introduce a global feature encoder, denoted as $\mathcal{F}_g(\cdot)$, to extract the global features $\mathbf{Z}_{g} \in \mathbb{R}^{N \times 1 \times D}$ for the entire image. This is based on the premise that the overall image encapsulates most of the essential information. However, it is crucial to pinpoint the specific location of the skin disease's origin. Moreover, global features often encompass background noise, which can degrade the prediction performance. Thus, while global features provide a broad context, their inherent noise necessitates complementary local analysis for precise disease localization and classification.

To address this issue, we propose exploring local visual concepts that potentially preserve more focused information about skin disease, thereby enhancing model decision-making. As illustrated in Figure \ref{framework}, we input the local visual concepts $\mathbf{V}$ into the local encoder $\mathcal{F}_l(\cdot)$ to extract local features $\mathbf{Z}_{l} \in \mathbb{R}^{N \times n \times D}$, where $n$ represents the total number of visual concepts. This method effectively filters out background noise and irrelevant objects, ensuring the model focuses on the most severely affected and informative pixels.

To efficiently leverage the complementary information between local concepts and global images, we propose a cross-attentive module where local concepts serve as ``query'' prompts to highlight the most salient areas within the global features. Specifically, the attention map $\mathbf{M} \in \mathbb{R}^{N \times n \times 1}$ acts as a soft mask between the local concept features $\mathbf{Z}_{l}$ and the global features $\mathbf{Z}_{g}$. This attention map is normalized using a global $\mathsf{softmax}$ function to capture the most relevant segments between the local and global features. This component further strengthens the model's ability to focus on the critical areas, thereby enhancing the overall precision and effectiveness of the diagnosis system.

Following the Transformer design, we define three variables—query, key, and value—as $\mathbf{Q} = \mathbf{Z}_{l}\mathbf{W}_{q}$, $\mathbf{K} = \mathbf{Z}_{g}\mathbf{W}_{k}$, and $\mathbf{V} = \mathbf{Z}_{g}\mathbf{W}_{v}$, where $\mathbf{W}_{q}$, $\mathbf{W}_{k}$, and $\mathbf{W}_{v}$ are the linear projection matrices in $\mathbb{R}^{D \times D}$ used to generate the query $\mathbf{Q}$, key $\mathbf{K}$, and value $\mathbf{V}$, respectively. These maps are then processed by each Transformer module to generate refined latent features, as described as follows:
\begin{equation}
\label{transformer equation}
\left\{
\begin{aligned}
& \mathbf{I} = \mathbf{Z}_{g} + \mathcal{F}_\mathrm{Drop}(\mathbf{M}\mathbf{V}\mathbf{W}_{proj}),\\
& \mathbf{O} = \mathbf{I} + \mathcal{F}_\mathrm{MLP}(\mathcal{F}_\mathrm{LN}(\mathbf{I})),
\end{aligned}
\right.
\end{equation}
where $\mathbf{W}_{proj} \in \mathbb{R}^{D \times D}$ are learnable projection matrices, $\mathcal{F}_\mathrm{LN}(\cdot)$ is the layer normalization function, $\mathcal{F}_\mathrm{Drop}(\cdot)$ is the dropout module, and $\mathcal{F}_\mathrm{MLP}(\cdot)$ is the multi-layer perceptron (MLP) module. The intermediate latent feature is denoted as $\mathbf{I}$. The output $\mathbf{O} \in \mathbb{R}^{N \times n \times D}$, which merges information from both local and global features, is more representative of the input image. To build the cross-attentive modules, we define $\mathbf{M} = \frac{\mathbf{Q}\mathbf{K}^{\top}}{\sqrt{n}}$. Additionally, multi-head attention is introduced to expand the capacity of the attention modules, allowing for a richer and more nuanced representation of the relevant features. This enhanced structure ensures a more effective integration and utilization of both local and global information.

\renewcommand{\arraystretch}{1.1}

\begin{table*}[h]
\centering
\caption{Class distribution for the experimental dataset. ``AC Dermatitis'' is short for Allergic Contact Dermatitis.}\vspace{-3mm}
\resizebox{0.75\textwidth}{!}{%
\begin{tabular}{l|llllll}
\Xhline{1pt}
\textbf{NTD}     & Buruli ulcer & Yaws          & Scabies   & Leprosy     & Mycetoma  & Total \\\hline
Patient \#  & 300          & 223           & 160       & 57          & 18        & 758   \\
Image \# & 787          & 375           & 391       & 135         & 43        & 1,731  \\ \hline
\textbf{SCIN}     & Eczema       & AC Dermatitis & Urticaria & Insect Bite & Psoriasis & Total \\ \hline
Patient \#  & 355          & 214           & 163       & 128         & 61        & 921   \\
Image \# & 801          & 467           & 338       & 290         & 126       & 2,022  \\ \Xhline{1pt}
\end{tabular}
}
\label{dataset_description}\vspace{-4mm}
\end{table*}

\subsection{Interpretable Skin Diagnosis}

Having obtained the refined latent features $\mathbf{O}$, it is essential to identify which parts contribute the most to future observations in the medical field. Typically, most datasets lack such detailed labels because annotating them is labor-intensive and usually requires professional expertise to ensure label quality in medical research. Inspired by the weakly supervised setting described by \cite{hu2024weakly}, we introduce the Class Activation Map (CAM) and develop multi-instance learning (MIL) loss for the final prediction, instead of relying on traditional classification loss.

In our framework, we utilize a classifier $\mathcal{F}_{c}(\cdot)$, as shown in Figure \ref{framework}, to transform the latent features $\mathbf{O}$ into $\mathbf{O}_{cam} \in \mathbb{R}^{N \times n \times C}$. Subsequently, a top-$k$ strategy is employed to identify the concepts that contribute the most, denoted $\mathbf{\hat{O}}_{cam} \in \mathbb{R}^{N \times k \times C}$. The final prediction is based on the average value of these top-$k$ concepts, as given by:
\begin{equation}
\label{final_pred}
    \mathbf{O}_\mathrm{pred} = \mathcal{F}_\mathrm{AVG}(\mathbf{\hat{O}}_\mathrm{cam}),
\end{equation}
where $\mathcal{F}_\mathrm{AVG}(\cdot)$ represents the average pooling function along the dimension $k$. The final loss is formulated as:
\begin{equation}
    \mathcal{L} = \mathcal{L}_\mathrm{MIL}(\mathbf{O}_\mathrm{pred}, \mathbf{y}_{i}),
\end{equation}
where $\mathcal{L}_\mathrm{MIL}(\cdot)$ denotes the multi-instance learning loss function. This approach allows us to effectively utilize the weakly supervised signals to identify and focus on the most critical parts of the input image, thereby enhancing the accuracy and reliability of the final medical diagnosis.

Equation \eqref{final_pred} illustrates that the final decision is based on the average value of the top-$k$ visual concepts, indicating that these $k$ concepts provide the most significant contributions. However, in the context of a skin disease diagnosis system, it is common to search for the final interpretation with one most important local visual concept. As depicted in Figure \ref{interpretable_1}, our framework first determines the most probable skin disease from top-$k$ concepts. Subsequently, the single concept that contributes the most is used as the basis for interpreting our model's decision. This ensures that the diagnosis is not only accurate but also interpretable, allowing medical professionals to understand which specific feature the model deems most indicative of the disease.

\renewcommand{\arraystretch}{1.08}
\begin{table*}[t]
\centering
\caption{Comparison of different methods on the MIND-the-SKIN dataset.}\scriptsize
\scalebox{1.5}{
\begin{tabular}{c|c|cccc}
\Xhline{1pt}
\textbf{Split Ratio} & \textbf{Method} & \textbf{Precision} & \textbf{Recall} & \textbf{F1} & \textbf{Accuracy} \\ \hline
\multirow{4}{*}{train/total = 0.1} & ResNet\cite{he2016deep} & 0.381 & 0.324 & 0.350 & 0.535 \\
                      & ViT\cite{dosovitskiy2020image}    & \textbf{0.393} & 0.364 & 0.378 & 0.578 \\
                      & Baseline(global-only) & 0.313 & 0.330 & 0.321 & 0.539 \\
                      & Ours    & 0.375 & \textbf{0.381} & \textbf{0.378} & \textbf{0.579} \\ \cline{1-6}
\multirow{4}{*}{train/total = 0.3} & ResNet\cite{he2016deep} & 0.575 & \textbf{0.538} & 0.556 & 0.721 \\
                      & ViT\cite{dosovitskiy2020image}    & 0.603 & 0.536 & 0.567 & 0.746 \\
                      & Baseline (global-only) & 0.606 & 0.495 & 0.545 & 0.743 \\
                      & Ours    & \textbf{0.643} & 0.509 & \textbf{0.568} & \textbf{0.750} \\ \cline{1-6}
\multirow{4}{*}{train/total = 0.5} & ResNet\cite{he2016deep} & 0.623 & 0.547 & 0.582 & 0.777 \\
                      & ViT\cite{dosovitskiy2020image}    & 0.554 & 0.567 & 0.560 & 0.760 \\
                      & Baseline (global-only) & 0.550 & 0.565 & 0.557 & 0.764 \\
                      & Ours    & \textbf{0.651} & \textbf{0.631} & \textbf{0.641} & \textbf{0.809} \\ \cline{1-6}
\multirow{4}{*}{train/total = 0.7} & ResNet\cite{he2016deep} & 0.740 & 0.588 & 0.655 & 0.798 \\
                      & ViT\cite{dosovitskiy2020image}    & 0.756 & 0.663 & 0.706 & 0.789 \\
                      & Baseline (global-only) & 0.738 & 0.671 & 0.702 & 0.816 \\
                      & Ours    & \textbf{0.805} & \textbf{0.768} & \textbf{0.786} & \textbf{0.847} \\ \cline{1-6}
\multirow{4}{*}{train/total = 0.9} & ResNet\cite{he2016deep} & 0.739 & 0.732 & 0.735 & 0.829 \\
                      & ViT\cite{dosovitskiy2020image}    & 0.824 & 0.775 & 0.799 & 0.853 \\
                      & Baseline (global-only) & 0.766 & 0.682 & 0.722 & 0.842 \\
                      & Ours    & \textbf{0.841} & \textbf{0.798} & \textbf{0.819} & \textbf{0.867} \\ \Xhline{1pt}
\end{tabular}}\vspace{-3mm}
\label{ntd_results}
\end{table*}

\begin{figure*}[t]
    \centering
    \includegraphics[width=1\linewidth]{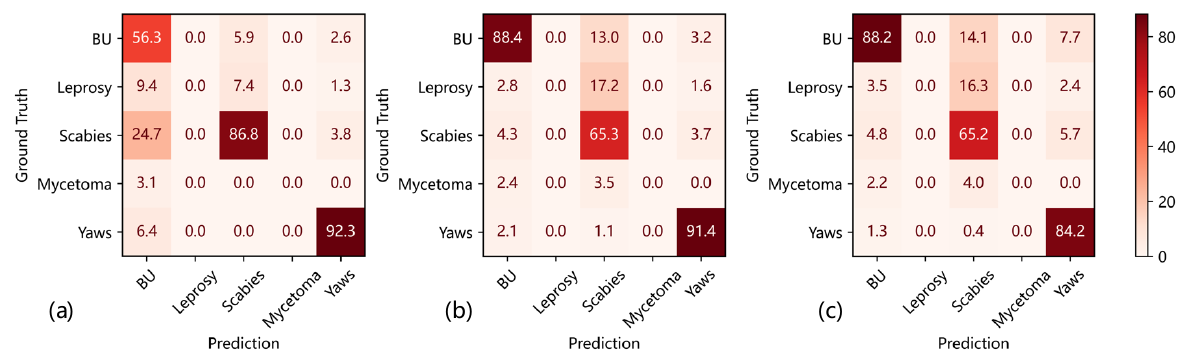}\vspace{-3mm}
    \caption{ Confusion matrix with different top-$k$ strategies on the MIND-the-SKIN dataset. \textbf{(a)} is the confusion matrix of top-1; \textbf{(b)} shows the confusion matrix of top-5; \textbf{(c)} represents the confusion matrix of top-15.}
    \label{confusion_matrix}\vspace{-3mm}
\end{figure*}

\section{Experiments}

\subsection{Datasets}

We evaluated our methods using two datasets: MIND-the-SKIN \cite{mindtheskin} and SCIN \cite{ward2024crowdsourcing}. The MIND-the-SKIN project aims to address challenges in Neglected Tropical Diseases (NTDs), a diverse group of skin conditions prevalent in impoverished tropical communities, affecting over 1 billion people. A crucial aspect of the project involves data collection in rural West Africa using portable devices and the development of AI-based diagnostic tools. For our evaluation, we utilized a subset of this dataset comprising 1,731 clinical photos representing five common NTD conditions: leprosy, Buruli ulcers, yaws, scabies, and mycetoma.

Existing skin disease datasets featuring clinical photos \cite{groh2021evaluating, daneshjou2022disparities, pacheco2020pad, yang2019self, giotis2015med} are pre-processed to center at lesions and have reduced background noise. In contrast, the SCIN dataset retains its original noise and more closely mirrors real-world inputs, making it a valuable resource for our investigation. The SCIN dataset was collected through a voluntary image donation platform from Google Search users in the United States, with each case including up to three images, all diagnosed by up to three dermatologists. This process results in a weighted skin condition label for each case. To ensure label accuracy, we selected the condition with the highest weight as the final label, excluding ambiguous cases where multiple conditions had equal probabilities. Additionally, to facilitate reliable evaluation and maintain consistency with the NTD dataset in terms of the number of conditions, we focused on the five largest classes. In our experiments, we used a random split by cases for both datasets to prevent data leakage. Dataset details and class distributions are provided in Table \ref{dataset_description}.

\subsection{Implementation Details}

We use the pretrained ViT \cite{dosovitskiy2020image} as the backbone to extract global features from each input image. The global feature extracted by ViT, combined with our top-$k$ mechanism, serves as the baseline for comparison. ViT produces a 768-dimensional feature map with a resolution of 14 $\times$ 14, which we pass through a linear layer to reduce the dimensionality to 256 for further refinement. For the local encoder, we extract visual concepts using our SAM variant, generating 256-dimensional vectors. These local features are refined with two convolutional layers, also outputting 256-dimensional vectors. To maintain consistent input, we fix the number of visual concepts at 30 for the NTD dataset and 25 for the SCIN dataset, based on the maximum number of captured concepts. For samples with fewer visual concepts, we use a random perturbation padding method to ensure consistency in input size. In the cross-attentive module, we implement a multi-head transformer with 8 heads, where each transformer module consists of 1 attention layer with a dimensionality of 256. The classification module includes 3 convolutional layers, with two dropout layers (dropout rate of 0.7) between them to regularize intermediate features.

For hyperparameter choice, we train the whole model with a learning rate of 1e-4 using the Adam optimizer, and the batch size is set as 128 for training and 1 for testing. Considering the overfitting problem, we train all model parts simultaneously with 30 epochs. We choose $k$ = 5 for the NTD dataset and 2 for the SCIN dataset. All experiments are conducted with one RTX 4090 GPU.

\begin{table*}[t]
\centering 
\caption{Comparison of different methods on the SCIN dataset.}
\scriptsize
\scalebox{1.52}{
\begin{tabular}{c|c|cccc}
\Xhline{1pt}
\textbf{Split Ratio} & \textbf{Method} & \textbf{Precision} & \textbf{Recall} & \textbf{F1} & \textbf{Accuracy} \\ \hline
\multirow{2}{*}{train/total = 0.1} & Baseline (global-only) & 0.196 & 0.204 & 0.200 & 0.385 \\
                      & Ours & \textbf{0.233} & \textbf{0.209} & \textbf{0.220} & \textbf{0.393} \\ \cline{1-6}
\multirow{2}{*}{train/total = 0.3} & Baseline (global-only) & 0.070 & 0.199 & 0.104 & 0.394 \\
                      & Ours & \textbf{0.190} & \textbf{0.221} & \textbf{0.204} & \textbf{0.395} \\ \cline{1-6}
\multirow{2}{*}{train/total = 0.5} & Baseline (global-only) & 0.145 & \textbf{0.209} & 0.171 & 0.395 \\
                      & Ours & \textbf{0.155} & 0.208 & \textbf{0.177} & \textbf{0.397} \\ \cline{1-6}
\multirow{2}{*}{train/total = 0.7} & Baseline (global-only) & 0.207 & 0.249 & 0.226 & \textbf{0.403} \\
                      & Ours & \textbf{0.275} & \textbf{0.255} & \textbf{0.265} & 0.394 \\ \cline{1-6}
\multirow{2}{*}{train/total = 0.9} & Baseline (global-only) & 0.325 & 0.307 & 0.316 & 0.436 \\
                      & Ours & \textbf{0.347} & \textbf{0.319} & \textbf{0.332} & \textbf{0.439} \\ \Xhline{1pt}
\end{tabular}}
\label{scin_results}\vspace{-1mm}
\end{table*}

\subsection{Comparison Results}

\begin{figure*}[t]
    \centering
    \includegraphics[width=1\linewidth]{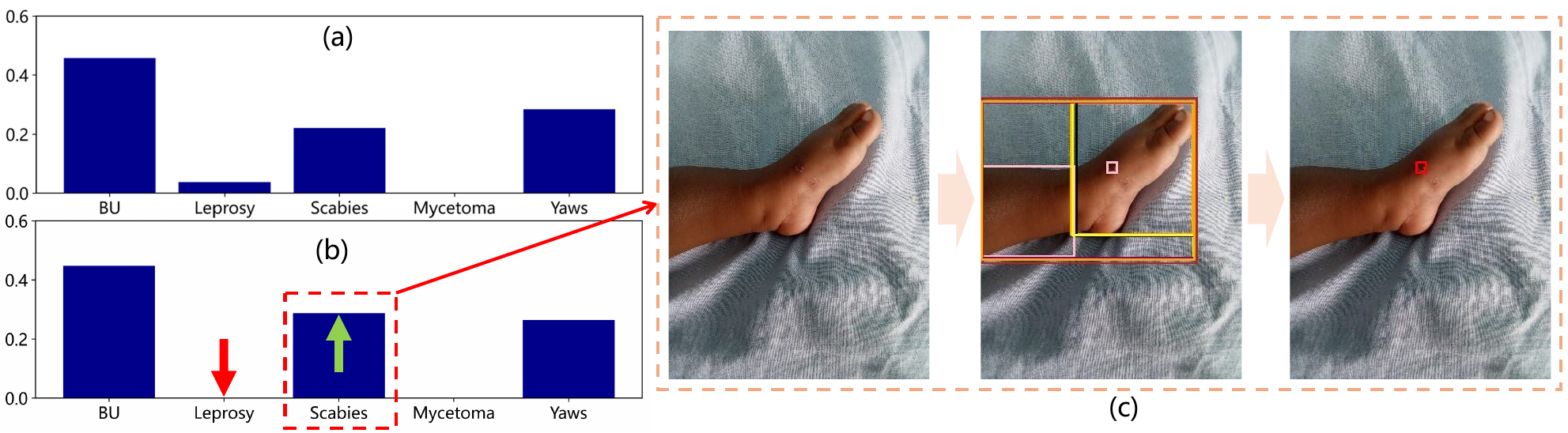}
    \vspace{-6mm}\caption{Performance comparison between baseline(global) and fusion(local + global) methods with the MIND-the-SKIN dataset. \textbf{(a)} shows the confidence for each condition by the baseline method; \textbf{(b)} the class-wise confidence by our fusion method; \textbf{(c)} demonstrates one ``Scabies'' example that is wrongly recognized by the baseline method and correctly predicted by our proposed fusion model.}\vspace{-3mm}
    \label{comparison_global_fusion}
\end{figure*}

We first tested our proposed method on the NTD dataset from the MIND-the-SKIN Project. As shown in Table \ref{ntd_results}, we conducted five random splits of cases into training and validation sets and compared our method with three other competitive methods. An immediate observation is that our method consistently outperforms the others in terms of classification accuracy for all splits, validating the effectiveness of leveraging both global and local features for skin lesion images. Notably, as the size of the training set increases, the margin of improvement gained by our method over the others also enlarges across the four metrics. This suggests that our method is more sensitive to new information than its competitors and can make full use of additional features to achieve superior performance.  Figure \ref{confusion_matrix} reports the confusion matrix when we select different $k$ and train/total = 0.5, which aims to show the category-specific accuracy. Interestingly, we notice our model fails to recognize ``Leprosy'' and ``Mycetoma''. The reason is that the lesion is not visually clear for those two diseases and they have the smallest sample sizes, thus easily misclassified. Figure \ref{comparison_global_fusion} indicates the comparison of each category in the MIND-the-SKIN dataset. It is observed that though our fusion model shows a little performance degradation in the ``Leprosy'' disease, it shows an obvious boost in the ``Scabies'' disease. On average, our proposed fusion method achieves better performance than the baseline method.

\begin{figure}
    \centering
\includegraphics[width=1\linewidth]{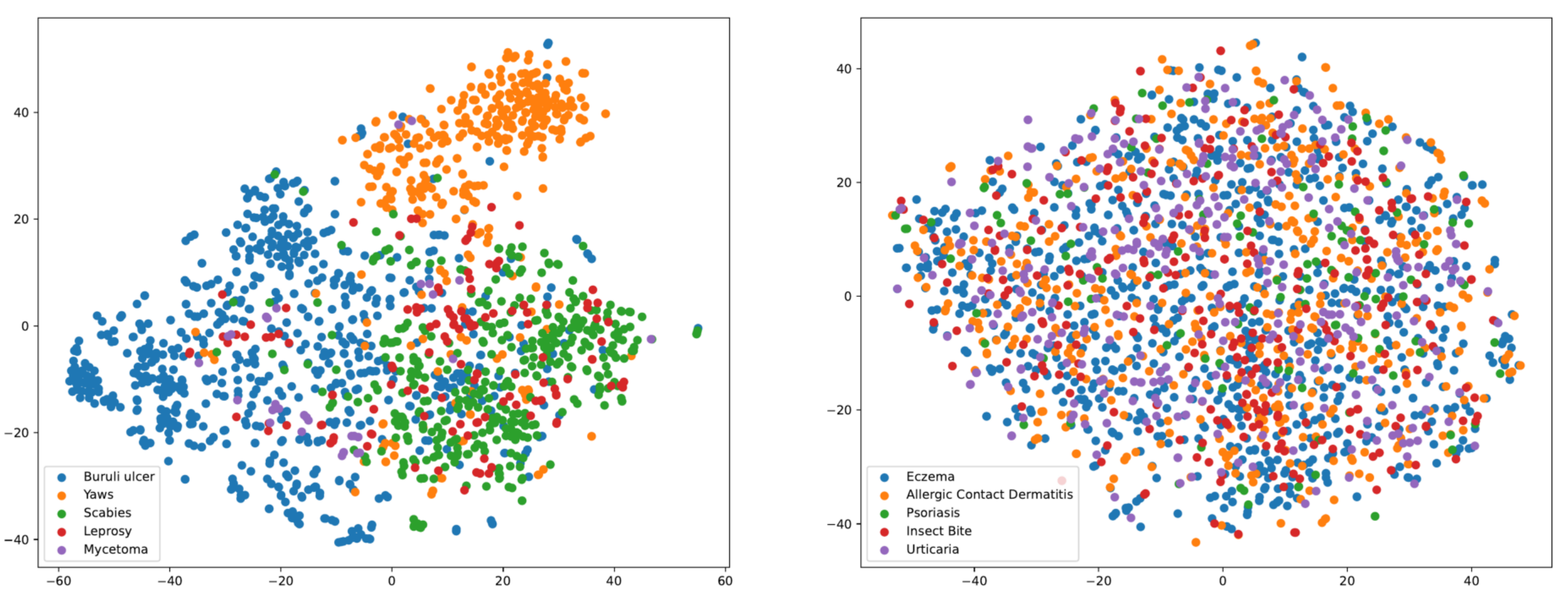}\vspace{-2mm}
    \caption{Visualization of data separability for the NTD (left) and SCIN (right) datasets after feature extraction by ViT.}\vspace{-6mm}
\label{fine_tuned_vs_original_sam2}
\end{figure}

Similarly, we performed five random splits for the SCIN dataset in Table \ref{scin_results}, as no official splits are available. We observed a similar trend in the SCIN dataset, where our proposed method generally outperforms the baseline. However, we also observed that the overall performance on the SCIN dataset was suboptimal. To the best of our knowledge, no prior studies have experimented with the SCIN dataset, leaving us no references for comparison. One possible explanation for the suboptimal performance is the difficulty in distinguishing between the conditions represented in the dataset, as shown by the t-SNE plots after feature extraction using ViT (Figure \ref{fine_tuned_vs_original_sam2}). Additionally, the dataset does not provide definitive labels for each image, but rather a list of potential labels with associated confidence scores, leading to label noise. Thus, while SCIN is the best available public dataset for our experiments, it remains challenging to use effectively, and its optimal usage is still unclear.

\begin{figure*}[!ht]
    \centering
    \includegraphics[width=1\linewidth]{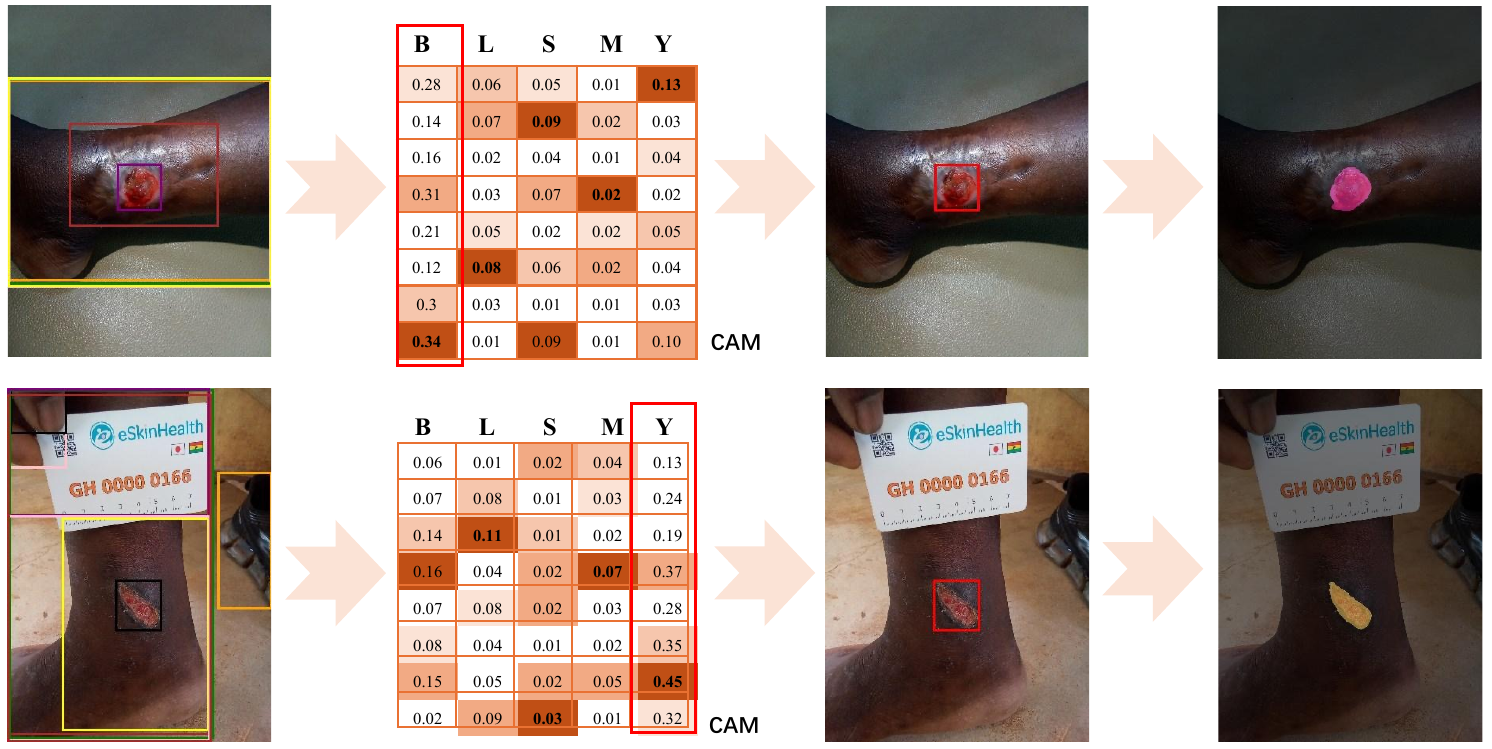}
\vspace{-6mm}\caption{Qualitative results of skin disease diagnosis in the MIND-the-SKIN dataset. The first row represents one example from the ``Buruli ulcer'' disease; The second row demonstrates the example from ``Yaws'' disease. Both samples select the most contributed visual concepts and prove the effectiveness of our proposed method.}
    \label{interpretable_1}\vspace{-4mm}
\end{figure*}

\subsection{Interpretable Results}
Figure \ref{interpretable_1} shows examples of “Buruli ulcer” and “Yaws,” where the diseased areas are easily identifiable on the skin. The distribution of the row scores in the CAM indicates that both cases have a high probability of the correct disease and low confidence in others. The column score distribution demonstrates that our model focuses more on the lesion part rather than the whole body. The selected bounding box (third column of the figure) and segmentation mask (fourth column) provide professionals with an intuitive yet reliable explanation for the prediction. Note that we report the final prediction using the top-5 visual concepts, and we select the top-1 from these to represent the visual concept that contributes most to the prediction, considering that multiple concepts might confuse the medical diagnosis.

\subsection{Ablation Study}

Table \ref{combine} indicates the performance comparison of different feature combinations on the MIND-the-SKIN dataset with train/total=0.5. The ``Local-only'' and ``Baseline'' denote the methods that only use local or global features individually for the final prediction. The ``Concatenation-$1$'' represents that we simply concatenate global feature to $n$ local features as $n$+1 dimensional feature $\mathbf{O'} \in \mathbb{R}^{N \times (n+1) \times D}$. ``Concatenation-$2$'' follows the feature fusion method in \cite{xu2020explainable} to combine local and global features in feature space as $\mathbf{O''} \in \mathbb{R}^{N \times n \times 2D}$. ``Average Sum'' conducts average pooling and sum operation on local and global features. The comparison results demonstrate that our cross-attention fusion method achieves better performance than other fusion strategies.

\renewcommand{\arraystretch}{1.2}
\begin{table}[t]
\centering
\caption{Comparison of different feature combining methods on the MIND-the-SKIN dataset with train/total=0.5.}\vspace{-1mm}
\scalebox{0.85}{
\begin{tabular}{c|cccc}
\Xhline{1pt}
\textbf{Methods}   & \textbf{Precision} & \textbf{Recall} & \textbf{F1} & \textbf{Accuracy} \\ \hline
Local-only              & 0.598              & 0.589           & 0.593       & 0.69              \\
Baseline (global-only)             & 0.550               & 0.565           & 0.557       & 0.764             \\
Concatenation-1 & 0.576              & 0.575           & 0.57        & 0.676             \\
Concatenation-2    & 0.555              & 0.571           & 0.562       & 0.671             \\
Average Sum        & 0.609              & 0.534           & 0.536       & 0.734             \\ 
Ours               & \textbf{0.651}     & \textbf{0.631}  & \textbf{0.641} & \textbf{0.809} \\     
\Xhline{1pt}
\end{tabular}}\vspace{-3mm}
\label{combine}
\end{table}

Table \ref{top_k} presents a performance comparison using different top-$k$ strategies. It is observed that the model achieves optimal performance when $k=5$. While fewer visual concepts are selected, there is a significant drop in precision, recall, and F1 score, even though the accuracy does not vary much. This can be attributed to the fact that accuracy is the metric used to select the best model, and the limited samples from the training dataset can lead to overfitting on this metric. Furthermore, when $k=15$, which is half the total number of concepts, there is a noticeable decline in all performance metrics. This decline occurs because including more visual concepts incorporates background noise from SAM into the final prediction, degrading performance.

\renewcommand{\arraystretch}{1.1}
\begin{table}[t]
\centering\vspace{-2mm}
\caption{Comparison of various $k$ on the MIND-the-SKIN dataset.}
\begin{tabular}{c|cccc}
\Xhline{1pt}
$k$ & \textbf{Precision} & \textbf{Recall} & \textbf{F1} & \textbf{Accuracy} \\ \hline
1          & 0.621              & 0.612           & 0.616       & 0.806             \\
2          & 0.615              & 0.624           & 0.619       & 0.807             \\
5          & \textbf{0.651}              & \textbf{0.631}           & \textbf{0.640}       & \textbf{0.809}             \\
6          & 0.647              & 0.634           & 0.640       & 0.808             \\
15         & 0.607              & 0.591           & 0.599       & 0.804             \\ \Xhline{1pt}
\end{tabular}\vspace{-1mm}
\label{top_k}
\end{table}

\section{Conclusion}

In this paper, we leverage the foundation AI model, SAM, to automatically segment visual skin images and propose a cross-attention model designed to harness complementary information between local visual concepts and global features in challenging clinical skin disease images. To effectively explain our model’s decision-making process, we integrate CAM and multi-instance learning to identify the most influential concepts, which are generated by SAM using stochastic text prompts. Our experiments demonstrate that the proposed method consistently outperforms competitive approaches across various metrics, underscoring the effectiveness of the dual-branch design. In addition, our method provides better interpretability, offering explainable predictions that enhance the reliability of AI-based diagnoses for medical professionals. This interpretability is crucial for building trust and improving diagnostic accessibility in future applications.

\clearpage

\balance
{\small
\bibliographystyle{ieee_fullname}
\bibliography{main}
}

\end{document}